\documentclass{article} 
\usepackage{nips15submit_e,times}
\usepackage{hyperref}
\usepackage{url}
\usepackage{graphicx}
\usepackage{amsmath}
\usepackage{amssymb}
\usepackage{subfig}

\title{Comparing Face Detection and Recognition Techniques}
\author{
Jyothi Korra\\
Computer Science and Engineering\\
Christ University Faculty of Engineering\\
Bangalore, India\\
\texttt{jyothi.korra@christuniversity.in} \\
}

\nipsfinalcopy % Uncomment for camera-ready version
\begin{document}
\maketitle

\begin{abstract}
This paper implements and compares different techniques for face detection and recognition. One is find where the face is located in the images that is face detection and second is face recognition that is identifying the person. We study three techniques in this paper: Face detection using self ­organizing map (SOM), Face recognition by projection and nearest neighbor and Face recognition using SVM.
\end{abstract}

\section{Face detection using SOM:}
A self­ organizing map (SOM) \cite{som,som2,som3} is a neural network based model. Each node maps a input vector to a scalar output using multiplicative weights. Typical SOM structure is a grid of neurons. The self­ organizing map is a dimensional reduction methods that is embed high dimensional vector to a low dimensional vector. The SOM trains models such that data points which are closer in high dimension also embed closer in lower dimension.
\subsection{Learning algorithm}
The training of the models starts from randomizing the weights of the neurons or initialized to principal components. Initializing the weights to principal components usually converges faster than random initialization \cite{somface}.
\subsection{Algorithm Steps}
\begin{itemize}
\item Identify closely similar images for training the map.
\item Convert the image into a grayscale image.
\item Split the image into 25 20x20 pixel images .
\item Initialize 25 5x5 SOMs with weight vectors representing n dimensions .The dimensions
correspond to those of mean grayness, FFT, texture histogram of the 20x20 images .
\item Initialize num of iterations = k, where k is the number of images used for training.
\item Initialize learning rate alpha = 0.9 (approx) .
\item Initialize dist = 4 , time constant = num of iterations/log(distance+1) , epoch = 1.
\item Pick a random training image.
\item For each 20x20 image segments of the training image do ,
a. calculate the mean grayness, FFT, grayscale histogram (with bin = 2 in our case) and initialize weight vector 'w' with these values.
b. compute the Euclidean distances between each neuron and $w$ for the SOM corresponding to the image segment $w_n­w$, $w_n$ being the weight vector of the neuron.
c. find the neuron which has the min Euclidean distance . Let this neuron be called winner
d. update the weights of the winner and all neurons within a distance 'dist' from the winner using the rule $$w_n = w_n+\alpha*(w­-w_n)$$
\item Repeat step 9 for other images by adjusting $$\alpha = \alpha * exp(­(epoch)/numofiter) $$ $$ dist = round(dist * exp(­(epoch)/timeconstant))$$.
\item Determine the winner neurons for all 25 SOMs representing different areas of the image.
\item Read the test image and preprocess the image and get the object boundaries within the image.
\item For each object found, do
1. Find test winner neurons of the 25 5x5 maps corresponding to 20x20 image segments of the test image.
2. Compare them with the winner neurons.
3. Allowing a threshold of 0 for each match, print exact match if the test image has 0 errors, i.e, all the winner neurons of the training set exactly match with the test winners.
\item If less than m dissimilarities result from the test neurons (for our case , m= 10) print close
match else print dissimilar.
\end{itemize}

\subsection{Face recognition using random projections and nearest neighbor}
Each face image is projected to lower dimension. Projections are done using Random projections and also Principle Component Analysis(PCA)\cite{pca}. Principal component analysis (PCA) is a technique used to reduce multidimensional datasets to lower dimensions for analysis. And in lower dimension perform nearest neighbor to recognition the face. PCA is also one of the primary techniques for identifying facial interest points \cite{pcaface,dg6,pcaface2,pcaface3}.
\subsubsection{Random Projection}
\begin{itemize}
\item Take a matrix with dimension as (dimesion of image) x (dimension to project) and fill this matrix with random values from Normal Distribution (0,1) .
\item Project all Training image to this space given by random matrix.
\item Given a test image project it to space given by random matrix. Using the projected dimension,
do nearest neighbour with projected training image.
\end{itemize}

\subsubsection{PCA}
\begin{itemize}
\item Take all training image as vectors. And find the covariance of these vectors.
\item Find the eigen vectors and value ues of $A'A$ where is matrix with each column as image
vector. Note actual PCA finds eigen values an vector of $AA'$. Since dimension is high than
number of data points doing this.
\item Project training image to eigen vectors found.
\item For test image, first project to eigen vectors and obtain projected vector and do nearest
neighbour using this dimension.
\end{itemize}

\subsubsection{Results}
Tested on two face data sets. 
\begin{itemize}
\item CalTech 101 face data set. Dataset is tagged using crowdsourcing approaches\cite{dg4}. Random Projection gives around 80\% accuracy and PCA gives around 69\%. Reason for high error rate in PCA is because eigen vector of A'A may not gives highest eigen values and vectors corresponds. It is an approximation for that.
\item IIT Kanpur face data set:
Random Projection gives around 92\% and PCA around 90\%. Reason for low error rate in
this data set is because background is constant. But in CalTech backgrounds are different so high error
rate.
\end{itemize}

The reason why PCA is giving bad results than Random projection may be because in PCA,
since dimension of image is very large computing eigen value and eigen vector for $AA'$ (where A is matrix, each column is an image) is not possible in general system. Because dimension of eigen vector of $AA'$ is dimension of image and number of eigen vectors of $AA'$ is also dimension of image. So, instead of calculating eigen vector of $AA'$, calculate eigen vector of $A'A$ and multiply eigen vectors of $A'A$ with A, which will give only n eigen vectors of $AA'$, where n is number of data points. And do projections in this eigen vectors. This eigen vectors need not corresponds to highest eigen values of $AA'$.

\subsection{SVM based algorithms}
The usability of raw SVMs\cite{svm} for the purpose of face recognition was investigated. The methods
tried out include:
\begin{itemize}
\item Raw SVMs with input points as entire images.
\item Sift points extracted and location of the first ten sift points used as the features.
\item Sift points extracted and the descriptors of the first ten sift points used as the features.
\item Sift points extracted and filtered through edge detectors. Then the descriptors of the top ten
sift points used as the features.
\item Implemented and tested existing uses of SVMs for the purpose of face recognition -
PCA­SVM and ICA­SVM
\end{itemize}
The usability of SVMs \cite{dg3} for other problems involving classification of images was investigated.
With specific focus on the problem of identification of displayed digits on Broadcast (Television) Quality Videos. A fast and online svm­ learning algorithm has been implemented based on the paper Fast Kernel classifiers with online and Active Learning.
\subsubsection{Results}
Even slight changes in posture leads to misclassification. Very bad for the purpose of face
recognition. Difference in background and lighting conditions lead to problems. But changes in posture are no more a problem. Comparable performance to method 2. Improvement in terms of translated and scaled images. Best among the four. But still is affected by differential lighting conditions(which leads to difference in the Sift interest points). Accuracy of around 60\% on Caltech face recognition data. PCA­SVM and ICA­SVM are both really good for the purpose of face recognition. Even in cases where face detection has not succeeded in localizing the face accurately. It shows good results even when the recognition task involves stuff like tilted faces, different expressions etc. ICA­SVM was found to be the better case when the task involves images of higher dimensions. \\
Accuracy of more than 80\% was obtained with leave one out error calculation on IITKanpur Face Recognition Dataset. The results found are very encouraging. A normal SVM (Soft margin) with quadratic polynomial kernel was found to do the task of classification of such digits with a very good percentage of accuracy(90\%). The test data involved data collected from Cricket videos across various series and across various spans of time. (Which involves varied fonts and varies aspect ration for the digits).
\section{Conclusion}
SVM is a very good tool for most specific tasks of object recognition\cite{dg1,dg2,dg5} in videos. To get the best results, the various combinations of Kernel types and hyper parameter values should be experimented. The selection of the training set is of special importance. If wrong set of training samples are selected, the performance will surely go down. It is important to normalize the data being given for training and for classification. Proper selection of classification features also helps in improving the performance of SVMs on such problems.\\
Possible set of features found are: SIFT descriptors of corner points, Inter point distance measures within the object, the projections of the difference along the basis vectors of the subspace generated by the difference of the images from their mean. (Eigen vectors of the Covariance Matrix), more generically, the projections of the image along any set of independent directions along the subspace of training samples.
{\small
\bibliographystyle{ieee}
\bibliography{egbib}
}

\end{document}